# 3-SAT Problem: A New Memetic-PSO Algorithm

Nasser Lotfi
Department of Computer Engineering
EMU University
Famagusta, North Cyprus
nasser.lotfi@emu.edu.tr

Jamshid Tamouk
Department of Computer Engineering
EMU University
Famagusta, North Cyprus
Jamshid.tamouk@cc.emu.edu.tr

Mina Farmanbar
Department of Computer Engineering
EMU University
Famagusta, North Cyprus
mina.farmanbar@emu.edu.tr

*Abstract*—**3-SAT problem is of great importance to many technical and scientific applications. This paper presents a new hybrid evolutionary algorithm for solving this satisfiability problem. 3-SAT problem has the huge search space and hence it is known as a NP-hard problem. So, deterministic approaches are not applicable in this context. Thereof, application of evolutionary processing approaches and especially PSO will be very effective for solving these kinds of problems. In this paper, we introduce a new evolutionary optimization technique based on PSO, Memetic algorithm and local search approaches. When some heuristics are mixed, their advantages are collected as well and we can reach to the better outcomes. Finally, we test our proposed algorithm over some benchmarks used by some another available algorithms. Obtained results show that our new method leads to the suitable results by the appropriate time. Thereby, it achieves a better result in compared with the existent approaches such as pure genetic algorithm and some verified types.**

## I. INTRODUCTION

3-SAT problem is of great importance to achieve higher performance in many applications. This paper presents a new hybrid evolutionary algorithm for solving this satisfiability problem. 3-SAT problem has the huge search space and it is a NP-hard problem [1]. Therefore, deterministic approaches are not recommended for optimizing of these functions with a large number of variables [2]. In contrast, an evolutionary approach such as PSO may be applied to solve these kinds of problems, effectively. There exist a few genetic algorithms for solving 3-SAT problem. The representation of a problem solution, encoding scheme, highly affects the speed of genetic algorithms. The primary difference amongst genetic algorithms is the chromosomal representation, Crossover scheme, mutation Scheme and Selection strategy. Evolutionary optimization algorithms mainly encode the value of variables as string of bits. But the reported results show that they alone cannot approach to optimal point sufficiently. Also these algorithms spend more time to get these results. The performance of an evolutionary algorithm is often sensitive to the quality of its initial population [2]. A suitable choice of the initial population may accelerate the convergence rate of evolutionary algorithms because, having an initial population with better fitness values, the number of generations required to get the final individuals, may reduce. Further, high diversity in the population inhibits early convergence to a locally optimal solution [2]. In our produced way we observe this rule and produce the initial particles intelligently. The initial population of particles is usually generated randomly. The "goodness" of the initial population depends both on the average fitness (that is, the objective function value) of individuals in the population and the diversity in the population [2]. Losing on either count tends to produce a poor evolutionary algorithm. As it is described in the future Sections, by creating an initial particles as intelligently, the convergence rate of our proposed algorithm is highly accelerated.

Previous genetic algorithms used the simple operators to produce new population that have weak diversity [2]. In our proposed algorithm we have used a suitable way to represent particles that have several advantages. Important one is that the count of population to reach the final population reduced, because the algorithm starts by the convenient initial particles. Finally, it achieves a better value in comparison with the existing approaches such as genetic algorithm.

The remaining parts of this paper are organized as follows: In Section 2, the 3-SAT problem is outlined. Section 3 presents a structure of PSO algorithms. In Section 4, the proposed algorithm based on PSO and Memetic algorithms are described. A practical evaluation of the proposed optimization algorithm is presented in Section 5. Finally, section 6 states the conclusion and future works.

## II. 3-SAT PROBLEM

In this section, description of the multivariable function is presented. The SAT problem is one of the most important optimization combinatorial problems because it is the first and one of the simplest of the many problems that have been proved to be NP-Complete [3]. A Boolean satisfiability problem (SAT) involves a Boolean formula F consisting of a set of Boolean variables $x_1, x_2, ..., x_n$. The formula F is in conjunctive normal form and it is a conjunction of m clauses $c_1, c_2, ..., c_m$. Each clause c, is a disjunction of one or more literals, where a literal is a variable $x_j$ or its negation. A formula F is satisfiable if there is a truth assignment to its variables satisfying every clause of the formula, otherwise the formula is unsatistiable. The goal is to determine a variable x assignment satisfying all clauses [4].

For example, in the formula below p1, p2, p3 and p4 are propositional variables. This formula is named CNF.

$$(p_1 \vee p_2 \vee \neg p_3) \wedge (\neg p_1 \vee p_2 \vee p_3) \wedge (\neg p_1 \vee \neg p_2 \vee p_3) \wedge (p_1 \vee \neg p_3 \vee p_4)$$

The class k-SAT contains all SAT instances where each clause contains exactly k distinct literals. While 2-SAT is solvable in polynomial time, k-SAT is NP-complete for $k \geq 3$ [5]. The SATs have many practical applications (e.g. in planning, in circuit design. in spin-glass model. in molecular biology ([6], [7], [8]) and especially many applications and research on the 3-SAT is reported. Many exact and heuristic algorithms have been introduced.

As described above in Section 1, 3-SAT optimization problem is a NP-hard problem which can be best solved by applying an evolutionary optimization approaches. In the following, we consider the PSO and Memetic algorithms and using them to solve this problem.

## III. PARTICLE SWARM OPTIMIZATION AND MEMETIC ALGORITHMS

Particle swarm optimization (PSO) [9] is a population based stochastic optimization technique developed by Dr. Eberhart and Dr. Kennedy in 1995, inspired by the social behavior of birds. The algorithm is very simple but powerful. A "swarm" is an apparently disorganized collection (population) of moving individuals that tend to cluster together while each individual seems to be moving in a random direction. We also use "swarm" to describe a certain family of social processes. The PSO approach utilizes a cooperative swarm of particles, where each particle represents a candidate solution, to explore the space of possible solutions to an optimization problem. Each particle is randomly or heuristically initialized and then allowed to 'fly' [9]. At each step of the optimization, each particle is allowed to evaluate its own fitness and the fitness of its neighboring particles. Each particle can keep track of its own solution, which resulted in the best fitness, as well as see the candidate solution for the best performing particle in its neighborhood. At each optimization step, indexed by t, each particle, indexed by i, adjusts its candidate solution (flies) according to (1) and Figure 1 [10].

$$\vec{v}_i(t+1) = \vec{v}_i(t) + \phi_1(\vec{x}_{i,p} - \vec{x}_i) + \phi_2(\vec{x}_{i,n} - \vec{x}_i)$$
$$\vec{x}_i(t+1) = \vec{x}_i(t) + \vec{v}_i(t+1) \qquad (1)$$

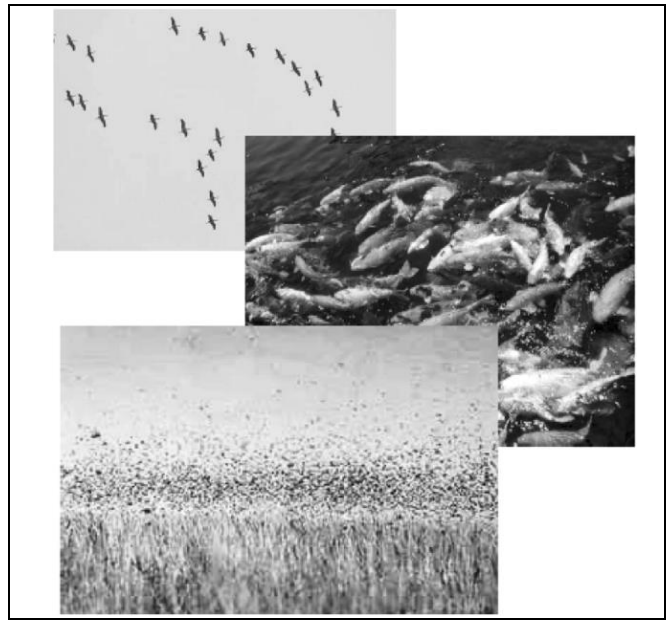

Figure1. Compute the particles's new location

First equation in (1) may be interpreted as the 'kinematic' equation of motion for one of the particles (test solution) of the swarm. The variables in the dynamical system of first equation are summarized in Table1 [10].

TABLE I. VARIABLES USED TO EVALUATE THE DYNAMICAL SWARM RESPONSE

| Parameter | Description |
|---|---|
| $\vec{v}_i$ | The particle velocity |
| $\vec{x}_i$ | The particle position (Test Solution) |
| t | Time |
| $\Phi_1$ | A uniform random variable usually distributed over [0,2] |
| $\Phi_2$ | A uniform random variable usually distributed over [0,2] |
| $\vec{x}_{i,p}$ | The particle's position (previous) that resulted in the best fitness so far |
| $\vec{x}_{i,n}$ | The neighborhood position that resulted in the best fitness so far |

Figure 2 shows the Algorithm pseudo code of PSO Generally.

```
I ) For each particle:
        Initialize particles.
II ) Do:
    a) For each particle:
        1) Calculate fitness value
        2) If the fitness value is better than the best Fitness
           value  (pBest) in history
        3) Set current value as the new pBest
      End
    b) For each particle:
        1) Find in the particle neighborhood, the particle With
           the best fitness
        2) Calculate particle velocity according to the
           Velocity equation
        3) Apply the velocity constriction
        4) Update particle position according to the
           Position equation
        5) Apply the position constriction
      End
    While maximum iterations or minimum error criteria is not attained.
```

Figure 2. The PSO Algorithm pseudo code.

The combination of Evolutionary Algorithms with Local Search Operators that work within the EA loop has been termed "Memetic Algorithms". Term also applies to EAs that use instance specific knowledge in operators. Local search is the searching of best solution among adjacent solutions that replace population members with better than. Pivot rule in the memetic algorithms have two types. At first type the search stopped as soon as a fitter neighbor is found (Greedy Ascent) and at second type the whole set of neighbors examined and the best neighbor found (Steepest Ascent). Figure 3 shows the pseudo code for local search [11].

```
Begin
/* given a starting solution i and a  neighborhood function*/
Set best =i ;
Set iteration =0;
Repeat until (depth condition is satisfied )  DO
  Set count =1;
  Repeat until (pivot rule is satisfied) DO
    Generate the next neighbor j ∈ n(i)
    Set count =count+1;
    IF (f(j) is better than f (best) THEN
      Set best =j;
    FI
  OD
  Set i=best
  Set iteration =iteration+1
OD
```

Figure 3. The local search pseudo code

It has been shown that the memetic algorithms are faster and more accurate than GAs on some problems, and are the "state of the art" on many problems. Another common approach would be to initialize population with solutions already known, or found by another technique (beware, performance may appear to drop at first if local optima on different landscapes do not coincide) [11].

IV. A NEW MEMETIC PSO TO SOLVE 3-SAT PROBLEM

To understand the algorithm, it is best to imagine a swarm of birds that are searching for food in a defined area - there is only one piece of food in this area. Initially, the birds don't know where the food is, but they know at each time how far the food is. Which strategy will the birds follow? Well, each bird will follow the one that is nearest to the food [8].

PSO adapts this behavior and searches for the best solution-vector in the search space. A single solution is called particle. Each particle has a fitness/cost value that is evaluated by the function to be minimized, and each particle has a velocity that directs the "flying" of the particles. The particles fly through the search space by following the optimum particles [8].

The algorithm is initialized with particles at random positions, and then it explores the search space to find better solutions. But in our proposed memetic-PSO algorithm, the initial population is not produce quite random. We must produce initial population with better quality than random type. In our proposed algorithm we combine PSO, Memetic and Local search algorithms to collect their advantages in a new algorithm. To attain this population we produce 1000 particle and then select the 100 better particles among them. Or in other words, we produce initial particles by heuristic to have better swarm. Each particle represented by the binary array inclusive just 0 and 1. Length of this array is equal to number of propositional variables. For a CNF with 32 variables, we can assume the length equal to 32. An example of the particle is given in Figure 4. In this particle, the values of first and last variables are TRUE and FALSE respectively.

```
1110011011001100 0110101110000010
```

Figure 4. A chromosome created by memetic approach

In the every iteration, each particle adjusts its velocity to follow two best solutions. The first is the cognitive part, where the particle follows its own best solution found so far. This is the solution that produces the lowest cost (has the highest fitness). This value is called pBest (particle best). The other best value is the current best solution of the swarm, i.e., the best solution by any particle in the swarm. This value is called gBest (global best). In the 3-SAT problem, we can not use the introduced PSO formulas, because the solutions or particles in this problem are binary. Hence we must use another form of PSO named by Binary PSO. In the binary PSO the formulas we can use are as following. Then, each particle adjusts its velocity and position with the equations below in Figure 5.

$$g(v_{id}) = \begin{cases} v_{max}, & v_{id} > v_{max} \\ v_{id}, & -v_{max} \leq v_{id} \leq v_{max} \\ -v_{max}, & v_{id} < -v_{max} \end{cases}$$

$$sig(v_{id}) = \frac{1}{1 + \exp(-v_{id})}$$

$$v_{id} = g(\omega v_{id} + c_1 R_{id}(p_{id} - x_{id}) + c_2 r_{id}(p_{gd} - x_{id}))$$

$$x_{id} = \begin{cases} 1, & rand < sig(v_{id}) \\ 0, & otehrwise \end{cases}$$

Figure 5. Velocity and position adjustent in binary PSO

In these formulas, $v_{id}$ and $x_{id}$ are the new velocity and position respectively, $P_{id}$ and $P_{gd}$ are Pbest and Gbest, $R_{id}$ and $r_{id}$ are even distributed random numbers in the interval [0, 1], and $c_1$ and $c_2$ are acceleration coefficients. The $c_1$ is the factor that influences the cognitive behavior, i.e., how much the particle will follow its own best solution and $c_2$ is the factor for social behavior, i.e., how much the particle will follow the swarm's best solution.

The algorithm can be written as follows in Figure 6 [8]:

1. Initialize each particle with a random velocity and random position.
2. Calculate the cost for each particle. If the current cost is lower than the best value so far, remember this position (pBest).
3. Choose the particle with the lowest cost of all particles. The position of this particle is gBest.
4. Calculate, for each particle, the new velocity and position according to the above equations.
5. Repeat steps 2-4 until maximum iteration or minimum error criteria is not attained.

Figure 6. Binary PSO Algorithm

This is a quite simple algorithm, but not sufficiently. In our new approach in order to produce high quality particles and having sufficient power, we add memetic approach again. After producing a population we use local search to each particle and improve that's quality. In other words, we use local search algorithm in the each iteration to replace particles by better neighbors. So each particle could improve itself and helps to speedy convergence to optimal point.

The quality of each particle is simply computed. Fitness value or quality of a particle is equal to the number of elements in CNF which the particle makes them TRUE or FALSE. Being TRUE or FALSE depends on our objective.

## V. EXPERIMENTAL RESULTS

In this section, the performance results and comparison of our proposed algorithm is presented. Our proposed algorithm is compared with the results of some existent algorithm [12, 13]. The comparison is made by applying our algorithm to the some famous CNFs presented in related papers. It is observed that the proposed algorithm results in better than other algorithms and it produces the better outcomes. However we don't compare our algorithm to another deterministic algorithm, because 3-SAT problem is NP-hard and Deterministic approaches are not applicable in this context. At first, we present the results of our proposed memetic PSO algorithm on random produced CNFs. Table below shows the obtained results.

TABLE II. RESULTS OVER RANDOM PRODUCED CNF'S

| Variable Number | Closure Number | Result | Validity | Generations |
|---|---|---|---|---|
| 36 | 12 | CNF is satisfiable | Valid | 100 |
| 33 | 65 | 7 Closure is not satisfiable | Valid | 200 |
| 62 | 74 | CNF is satisfiable | Valid | 120 |
| 100 | 100 | CNF is satisfiable | Valid | 150 |
| 80 | 50 | CNF is satisfiable | Valid | 80 |
| 50 | 50 | 1 Closure is not satisfiable | Valid | 200 |
| 93 | 77 | CNF is satisfiable | Valid | 134 |
| 83 | 32 | CNF is Satisfiable | Valid | 236 |
| 35 | 59 | CNF is Satisfiable | Valid | 176 |
| 43 | 90 | 9 Closure is not satisfiable | Valid | 200 |
| 26 | 79 | 3 Closure is not satisfiable | Valid | 200 |
| 88 | 57 | CNF is Satisfiable | Valid | 109 |
| 91 | 92 | CNF is Satisfiable | Valid | 167 |
| 98 | 56 | 1 Closure is not satisfiable | Valid | 200 |
| 45 | 78 | CNF is Satisfiable | Valid | 111 |
| 78 | 100 | CNF is Satisfiable | Valid | 136 |

Here we consider the sample CNF generated randomly with 100 variables and 100 Closures. Figure 7 shows the first population generated by memetic algorithm that's including the better particles. Variation between particles can be seen.

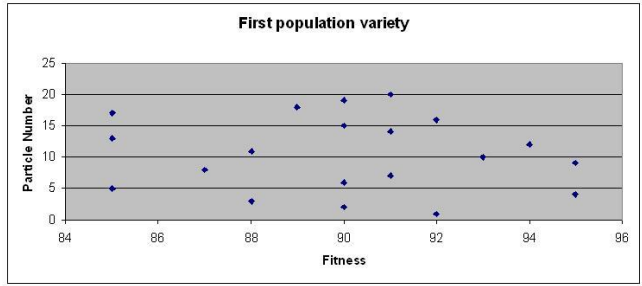

Figure 7. First population generated by memetic algorithm

The evolution of the chromosomes, while applying our proposed evolutionary algorithm on the mentioned example, is shown below in Figure 8. We can see that the fitness of best particle is gradually improved generation by generation.

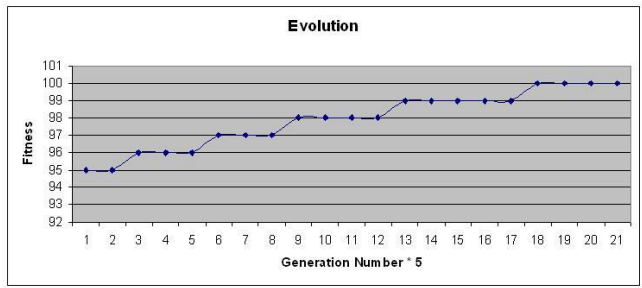

Figure 8. Evolution of particles

Also, in order to demonstrate the stability of the results obtained in the above example, the results obtained by twenty runs of the algorithm are compared in Figure 9. We can see that all 100 closures are satisfied in all 20 runs.

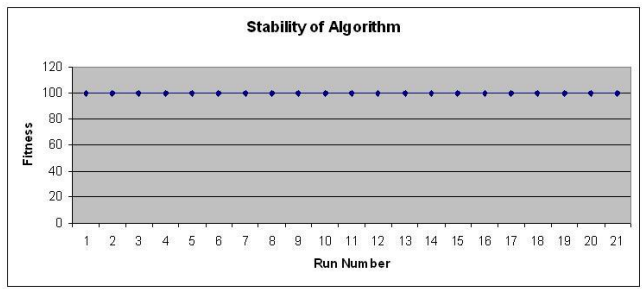

Figure 9. Best fitness obtained in 100 generations and 20 runs

We continue our evaluating using two existent well known algorithms to solve this problem [12, 13].

At first, we evaluate the performance of our proposed algorithm on several classes of satisfiable and unsatisfiable benchmark instances and compare it with GASAT [12] and with WALKSAT [14], one of the well-known incomplete algorithms for SAT, and with UNIWALK [15], the best up-to-now incomplete randomized solver presented to the SAT competitions [12]. Two classes of instances are used: structured and random instances. Structured instances are aim-100-1_6-yes1-4 (100 variables and 160 clauses), aim-100-2_0-yes1-3 (100 variables and 200 clauses), math25.shuffled (588 variables and 1968 clauses), math26.shuffled (744 variables and 2464 clauses), color-15-4 (900 variables and 45675 clauses), color-22-5 (2420 variables and 272129 clauses), g125.18 (2250 variables and 70163 clauses) and g250.29 (7250 variables and 454622 clauses).

Also, the random instances are glassy-v399-s1069116088 (399 variables and 1862 clauses), glassy-v450-s325799114 (450 variables and 2100 clauses), f1000 (1000 variables and 4250 clauses) and f2000 (2000 variables and 8500 clauses) [12]. Two criterions are used to evaluation and comparison. First one is the success rate (%) which is the number of successful runs divided by the total number of runs. The second criterion is the average running time in second. We have tried to use same computer and hardware for running [12]. Tables below show the comparison between these four algorithms. If no assignment is found then the best number of false clauses is written between parentheses.

TABLE III. STRUCTURED INSTANCES

| Benchmarks | Our Algorithm | GASAT | WALKSAT | UNITWALK |
|---|---|---|---|---|
| aim-100-1_6-yes1-4 | 100% <br> 27.19 | 10% <br> 84.53 | (1 clause) | 100% <br> 0.006 |
| aim-100-2_0-yes1-3 | 100% <br> 14.32 | 100% <br> 20.86 | (1 clause) | 100% <br> 0.0019 |
| math25.shuffled | (3 clauses) | (3 clauses) | (3 clauses) | (8 clauses) |
| math26.shuffled | (2 clauses) | (2 clauses) | (2 clauses) | (8 clauses) |
| color-15-4 | 100% <br> 358.43 | 100% <br> 479.248 | (7 clauses) | (16 clauses) |
| color-22-5 | (5 clauses) | (5 clauses) | (41 clauses) | (51 clauses) |
| g125.18 | 100% <br> 281.455 | 100% <br> 378.660 | (2 clauses) | (19 clauses) |
| g250.29 | (45 clauses) | (57 clauses) | (34 clauses) | (57 clauses) |

TABLE IV. RANDOM INSTANCES

| Benchmarks | Our Algorithm | GASAT | WALKSAT | UNITWALK |
|---|---|---|---|---|
| glassy-v399-s1069116088 | (5 clauses) | (5 clauses) | (5 clauses) | (17 clauses) |
| glassy-v450-s325799114 | (10 clauses) | (8 clauses) | (9 clauses) | (22 clauses) |
| F1000 | 100% <br> 34.45 | 100% <br> 227.649 | 100% <br> 9.634 | 100% <br> 1.091 |
| F2000 | 100% <br> 19.94 | (6 clauses) | 100% <br> 21.853 | 100% <br> 17.169 |

As we can see in tables, our proposed algorithm works better than others in overall and is more efficient from the performance view.

VI. CONCLUSIONS AND FUTURE WORKS

3-SAT problem is NP-hard and can be considered as an optimization problem. To solve this NP-hard problem, non-deterministic approaches such as evolutionary algorithms are quite effective.

Values of propositions can be best encoded as a binary array. The objective of evolutionary algorithms can be to maximize the number of valid DNF elements in CNF. In this way, the fitness of each particle in a population depends on the value of DNF elements. We used PSO approach based on memetic algorithms to solve this problem that is better than existent approaches.

The other kind of this problem is multi objective SAT problem that's more important. Multi-objective optimization problems consist of several objectives that are necessary to be handled simultaneously. Such problems arise in many applications, where two or more, sometimes competing and/or incommensurable, objective functions have to be minimized concurrently. It's possible to use evolutionary approaches to solve such problems [10].
Multivariable SAT problem can be defined in the form of multi-objective optimization problem. In this form, we deal with m formulas, each representing a different objective. The goal is to satisfy the maximum number of clauses in each formula. For solving this problem, we can extend our proposed memtic PSO to the multi-objective problems solver form. Hence, the set of non-dominated solutions must be found for this kind of problem.